\documentclass[11pt,a4paper]{article}
\usepackage[hyperref]{naaclhlt2019}
\usepackage{times}
\usepackage{latexsym}

\usepackage{url}

\aclfinalcopy %
\def\aclpaperid{723} %

\usepackage{float}
\usepackage{subcaption}
\usepackage{graphicx}
\usepackage{multirow}
\usepackage{float}
\usepackage{multicol}

\usepackage{url}            %
\usepackage{booktabs}       %
\usepackage{amsfonts}       %
\usepackage{nicefrac}       %
\usepackage{microtype}      %
\usepackage{multirow}
\usepackage{subcaption}
\usepackage{enumitem}
\usepackage{wrapfig}

\definecolor{Red}{rgb}{1,0, 0}
\definecolor{Green}{rgb}{0,1, 0}
\definecolor{Blue}{rgb}{0,0, 1}
\definecolor{Magenta}{rgb}{1,0,1}
\definecolor{BurntOrange}{rgb}{0.8,0.33, 0}

\title{VQD: Visual Query Detection in Natural Scenes}
\author{Manoj Acharya$^1$ \qquad  Karan Jariwala$^1$ \qquad  Christopher Kanan$^{1,2,3}$\\$^1$Rochester Institute of Technology\qquad $^2$PAIGE \qquad $^3$Cornell Tech\\
{\tt\small \{ma7583, kkj1811, kanan\}@rit.edu}}

\date{}
\begin{document}
\maketitle
\begin{abstract}
We propose Visual Query Detection (VQD), a new visual grounding task. In VQD, a system is guided by natural language to localize a \emph{variable} number of objects in an image. VQD is related to visual referring expression recognition, where the task is to localize only \emph{one} object. We describe the first dataset for VQD and we propose baseline algorithms that demonstrate the difficulty of the task compared to referring expression recognition.  %
\end{abstract}

\section{Introduction}

\begin{figure}[t!]
\centering
\footnotesize
\includegraphics[width=0.49\textwidth, ]{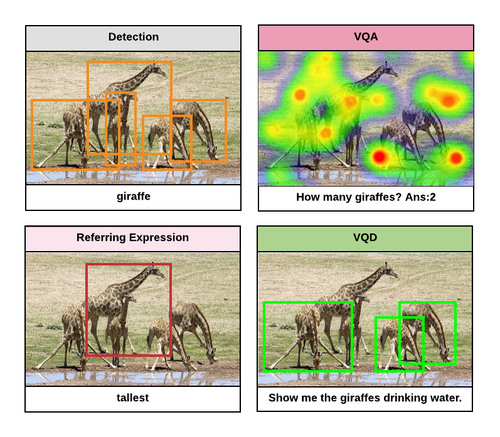}
\caption{Unlike VQD, object detection cannot deal with attributes and relations. In VQA, often algorithms produce the right answers due to dataset bias without `looking' at relevant image regions. RER datasets have short and often ambiguous prompts, and by having only a single box as an output, they make it easier to exploit dataset biases. VQD requires goal-directed object detection and outputting a variable number of boxes that answer a query.  \label{fig:overview-fig}}
\end{figure}

In computer vision, object detection is the task of identifying all objects from a specific closed-set of pre-defined classes by putting a bounding box around each object present in an image, e.g., in the widely used COCO dataset there are 80 object categories and an algorithm must put a box around all instances of each object present in an image~\cite{lin2014microsoft}. Recent deep learning based models have significantly advanced the state-of-the-art in object detection~\cite{ren2015faster}; however, many applications demand more nuanced detection of objects with specific attributes or objects in relation to each other.  Here, we study goal-directed object detection, where the set of possible valid objects is far greater than in the typical object detection problem. Specifically, we introduce the Visual Query Detection (VQD) task (see Fig.~\ref{fig:overview-fig}). In VQD, a system is given a query in natural language and an image and it must produce $0$--$N$ boxes that satisfy that query. VQD has numerous applications, ranging from image retrieval to robotics. 

VQD is related to the visual referring expression recognition (RER) task~\cite{kazemzadeh2014referitgame}; however, in RER every image has only a \emph{single} correct box. In contrast, in VQD there could be no valid outputs for a query or multiple valid outputs, making the task both harder and more useful. As discussed later, existing RER datasets have multiple annotation problems and have significant language bias problems. VQD is also related to Visual Question Answering (VQA), where the task is to answer questions about images in natural language~\cite{malinowski2014multi,antol2015vqa}. The key difference is that in VQD the algorithm must generate image bounding boxes that satisfy the query, making it less prone to the forms of bias that plague VQA datasets.

\textbf{We make the following contributions:}
\begin{enumerate}[noitemsep,nolistsep]
\item We describe the first dataset for VQD, which will be publicly released.
\item We evaluate multiple baselines on our VQD dataset.
\end{enumerate}

\section{Related work}

Over the past few years, a large amount of work has been done at the intersection of computer vision and natural language understanding, including visual madlibs~\cite{yu2015visual,tommasi2018combining},  captioning~\cite{farhadi2010every,kulkarni2013babytalk,johnson2016densecap,liu2018survey}, and image retrieval~\cite{wan2014deep,li2016socializing}. For VQD, the most related tasks are VQA and RER, which we review in detail.

\subsection{Visual Question Answering}

VQA systems take in an image and open-ended natural language question and then generate a text-based answer~\cite{antol2015vqa,goyal2017making,acharya2019tallyqa,kafle2018dvqa}. Many VQA datasets have been created. However, initial datasets, e.g., VQAv1~\cite{antol2015vqa} and COCO-QA~\cite{ren2015image}, exhibited significant language bias in which many questions could be answered correctly without looking at the image, e.g., for VQAv1 it was possible to achieve 50\% accuracy using language alone~\cite{kafle2016}. To address the bias issue, the VQAv2 dataset was created with a more balanced distribution for each possible answer to make algorithms analyze the image~\cite{goyal2017making}, but it still had bias in the kinds of questions asked, with some questions being scarce, e.g., reasoning questions. Synthetic datasets such as the CLEVR dataset \cite{johnson2017clevr} addressed this by being synthetically generated to emphasize hard reasoning questions that are rare in VQAv1 and VQAv2. The TDIUC dataset addresses bias using both synthetically generated and human gathered questions about natural images, with performance evaluated for 12 kinds of questions~\cite{kafle2017-iccv}. While the state-of-the-art has rapidly increased on both synthetic and natural image VQA datasets, many models do not generalize across datasets~\cite{shrestha2019answer}.

\subsection{Referring Expression Recognition}

Unlike VQA, RER algorithms must produce evidence to justify their outputs. A RER algorithm outputs a box around the image location matching the input string, making it easier to tell if an algorithm is behaving correctly. The RefCOCO and RefCOCO+ datasets for RER were collected from the two-player `ReferIt' Game~\cite{kazemzadeh2014referitgame}. The first player is asked to describe an outlined object and the second player has to correctly localize it from player one's description. The test datasets are futher split into the `testA' and `testB' splits. The split `testA' contains object categories sampled randomly to be close to the original data distribution, while `testB' contains objects sampled from the most frequent object categories, excluding categories such as `sky', `sand', `floor', etc. Since, there is a time limit on the game, the descriptions are short, e.g.,  `guy in a yellow t-shirt,' `pink,' etc.

Instead of playing a timed game, to create the RefCOCOg dataset for RER, one set of Amazon Mechanical Turk (AMT) users were asked to generate a description for a marked object in an image and other users marked the region corresponding to the description~\cite{Mao2016GenerationAC}. This resulted in more descriptive prompts compared to RefCOCO and RefCOCO+.

The Visual7W dataset for VQA includes a `pointing' task that is closely related to RER~\cite{visual7w}. Pointing questions require choosing which box of the four \emph{given} boxes correctly answered a query. Systems did not generate their own boxes, and there is always one correct box. 

\citet{cirik2018visual} showed that RER datasets suffer from biases caused by their dataset collection procedure. For RefCOCOg, they found that randomly permuting the word in the referring expression caused only about a 5\% drop in performance, suggesting that instead of relying on language structure, systems may be using some hidden correlations in language. They further showed that an image only model that ignores the referring expression yielded a precision of 71.2\% for top-2  best predictions. They also found that predicting the object category given the image region produced an accuracy of 84.2\% for top-2 best predictions. By having $0$--$N$ boxes, VQD is harder for an image-only model to perform well.

\section{The VQD 1.0 Dataset (VQDv1)}

\begin{figure*}[t]
\centering
\footnotesize
    \captionsetup[subfigure]{justification=centering}   
    \begin{subfigure}[t]{0.45\textwidth}
		\includegraphics[width=\textwidth, ]{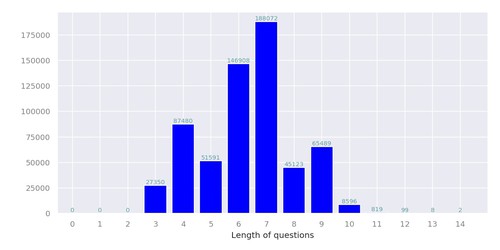}
        \caption{Question length distribution. \label{distr:c} }
    \end{subfigure}
    \hfill    
    \begin{subfigure}[t]{0.45\textwidth}
		\includegraphics[width=\textwidth, ]{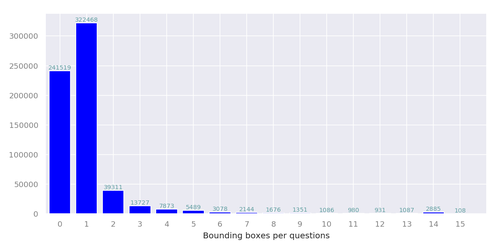}
        \caption{Bounding boxes per question distribution. \label{distr:b} }
    \end{subfigure}
        \begin{subfigure}[t]{0.60\textwidth}
		\includegraphics[width=\textwidth, ]{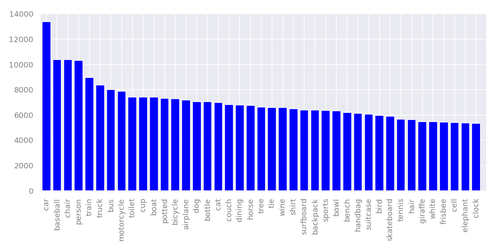}
        \caption{Top-40 Object category distribution. \label{distr:a} }
    \end{subfigure}
\caption{ Distribution statistics for the VQD dataset.\label{fig:distribution}}
\end{figure*}

\begin{figure*}[h!]
\footnotesize
    \captionsetup[subfigure]{justification=centering}   
    \begin{subfigure}[t]{0.32\textwidth}
		\includegraphics[width=\textwidth, ]{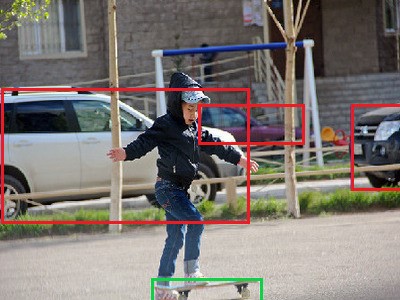}
        \caption{Show the car in the picture. (red) \\ Where is the skateboard in the image? (green) \label{subfig:a} }
    \end{subfigure}
    \hfill    
     \begin{subfigure}[t]{0.32\textwidth}
		\includegraphics[width=\textwidth, ]{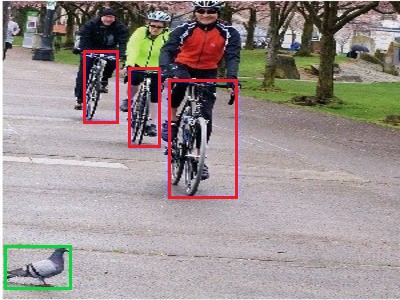}
        \caption{ Show me the bicycle. (red) \\ Where is the bird? (green)\label{subfig:b}}
    \end{subfigure}   
        \hfill    
     \begin{subfigure}[t]{0.32\textwidth}
		\includegraphics[width=\textwidth, ]{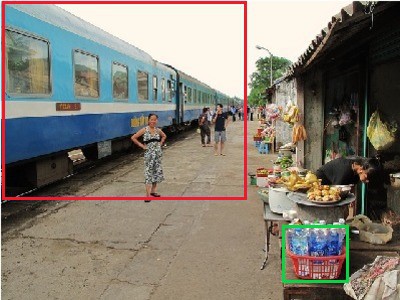}
        \caption{ Which train is blue in color? (red) \\ Show me the red basket. (green) \label{subfig:c}}
    \end{subfigure}   
        \hfill
     \begin{subfigure}[t]{0.32\textwidth}
		\includegraphics[width=\textwidth, ]{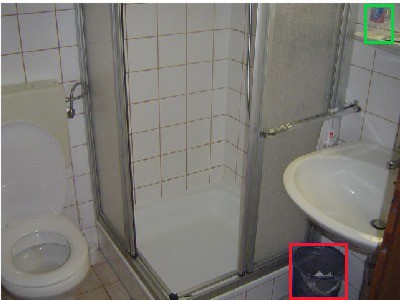}
        \caption{ Which bin is under the sink? (red) \\ Which bottle is on top of shelf? (green) \label{subfig:d}}
    \end{subfigure}   
    \hfill
     \begin{subfigure}[t]{0.32\textwidth}
		\includegraphics[width=\textwidth, ]{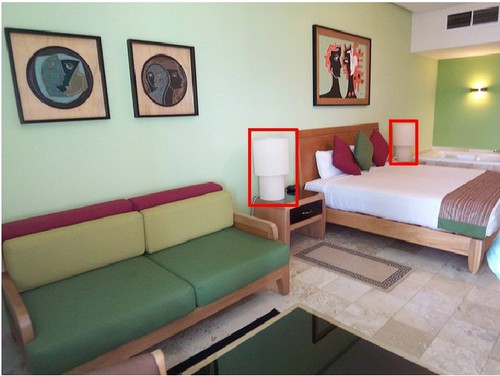}
        \caption{ Show the lamp beside bed in the image. (red) \label{subfig:e}}
    \end{subfigure}   
        \hfill
     \begin{subfigure}[t]{0.32\textwidth}
		\includegraphics[width=\textwidth, ]{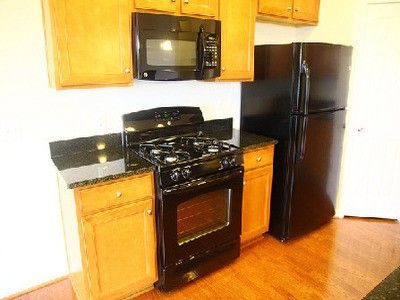}
        \caption{ Where is the sink in the picture? \\ Where is the toaster in the image? \label{subfig:f}}
    \end{subfigure}   
   
\caption{Example query-detection pairs from the VQD dataset. Counter context questions that do not have a bounding box as an answer are generated in such a way that they are still relevant to the scene context. For example, in Fig. [\ref{subfig:f}] both questions pertain to the context `kitchen'. \label{fig:examples} }
\end{figure*}

We created VQDv1, the first dataset for VQD. VQDv1 is created synthetically using annotations from Visual Genome (VG), COCO, and COCO Panoptic. While this limits variety, it helps combat some kinds of bias and serves as an initial version of the dataset. VQDv1 has three distinct query categories: 
\begin{enumerate}[noitemsep,nolistsep]
    \item Object Presence (e.g., `Show the dog in the image')
    \item Color Reasoning (e.g., `Which plate is white in color?')
    \item Positional Reasoning (e.g., `Show the cylinder behind the girl in the picture')
\end{enumerate}
The number of queries per type are given in the Table~\ref{table:vqd_categories}. The dataset statistics and example images and  are shown in Fig.~\ref{fig:distribution} and Fig.~\ref{fig:examples}, respectively. We show statistics for VQDv1 compared to RER datasets in Table~\ref{table:prevdataset}.

\begin{table}[t]
\footnotesize    %
\centering
\caption{VQDv1 Query Types}
\begin{tabular}{lr}
Type & \# Questions \\
\midrule
Simple & 391,628 \\
Color &  172,005 \\
Positional & 57,904 \\
\midrule
\textbf{Total} & 621,537 \\
\bottomrule
\label{table:vqd_categories}
\end{tabular}
\end{table}

\begin{table}[t]
\footnotesize    %
\caption{VQDv1 compared to RER datasets. \label{table:prevdataset} }
\centering
\begin{tabular}{lrr}
Dataset & \# Images & \# Questions \\
\midrule
RefCOCO & 19,994 & 142,209 \\
RefCOCO+ & 19,992 & 141,564 \\
RefCOCOg & 26,711 & 85,474 \\
VQDv1 & 123,287 & 621,537  \\
\bottomrule
\end{tabular}

\end{table}

All images in VQDv1 are from COCO. The ground truth bounding box annotations are derived from the COCO Panoptic annotations dataset~\cite{Kirillov2018PanopticS}. The questions are generated using multiple templates for each question type, which is an approach that has been used in earlier work for VQA~\cite{kafle2017-iccv,kafle2017b}.
The query objects and their attributes are extracted by integrating the annotations from images that have both COCO and VG annotations. COCO annotations are focused on objects, while VG also has attribute and relationship information, e.g., size, color, and actions for scene objects.

\subsection{Object Presence}
Object presence questions require an algorithm to determine all instances of an object in an image without any relationship or positional attributes, for example, `Show me the horse in the image' or `Where is the chair?' We  use all of the COCO `things' labels and half of the COCO `stuff' labels to generate these questions, making this task test the same capabilities as conventional object detection. We filter some `stuff' categories that do not have well defined  bounding boxes  such as `water-other', `floor-stone', etc. We use multiple templates to create variety, e.g., `Show the \textless object\textgreater~in the image', `Where are the \textless object\textgreater~in the picture?' etc.

\subsection{Color Reasoning}
Color questions test the presence of objects modified by color attributes, e.g., `Show me the cat which is grey in color' or `Which blanket is blue in color?' Since, COCO has only object annotations, color attributes are derived from VG's attribute annotations. We align every VG image annotation with COCO annotations to obtain (object, color) annotations for each bounding box. When multiple color attributes for an object are present, the object is assigned a single color from that attribute set. %

\subsection{Positional Reasoning}
Positional reasoning questions test the location of objects with respect to other objects, e.g., `Show the building behind horses', `Which people are in front of the lighthouse?', and `Show the rhino behind elephant.' We again use VG's relationship and attribute annotations to create these questions.

\subsection{Generating Counter-Concept Questions}

Counter-concept questions have no valid boxes as outputs, and we endeavor to create hard counter-concept questions for each category. We ask `Show me the zebra' only if there is a similar animal present (e.g., a cow), which was done by using COCO's super-categories. Likewise, `Show me the donut that is brown in color' is only asked if a brown donut does not exist in the image.

\section{Experiments}
\begin{table*}[t!]
\footnotesize
\centering
\caption{Results on RER datasets and two versions of our VQD dataset. The `1 Obj' version is trained and evaluated on queries with only a single box, analogous to RER, and the $0$--$N$ version contains the entire VQD dataset. All models use the same object proposals and visual features.}
\begin{tabular}{l|rrr|rrr|rr|rr}
& \multicolumn{3}{c}{RefCOCO} & \multicolumn{3}{|c|}{RefCOCO+} & \multicolumn{2}{|c}{RefCOCOg} & \multicolumn{2}{|c}{VQDv1}\\
& val & testA & testB & val & testA & testB & val & test & 1 Obj. & $0$--$N$ Obj. \\
\midrule
DETECT   & 38.63 & 37.82 &  38.32 & 38.85  & 37.85 & 38.98 & 50.13 & 50.03 & 30.44 & 26.94 \\
RANDOM   & 16.51 & 14.30 &  19.81 & 16.67  & 14.10 & 20.45 & 19.87 & 19.76 & 9.77 & 2.38  \\
Query-Blind & 33.95 & 37.28 & 31.58 & 34.06  & 37.34 & 32.46 & 39.79 & 23.34 & 23.34 & 6.80 \\
Vision+Query& 69.41 & \textbf{75.52} & \textbf{65.28} & \textbf{59.83}  & \textbf{65.21} & \textbf{53.02} & \textbf{62.52} & \textbf{62.06} & \textbf{37.55} & \textbf{31.03} \\
\midrule
\textbf{S}LR & \textbf{69.48} & 73.71 & 64.96 & 55.71 & 60.74 & 48.80 & 60.21 & 59.63 & --& --\\
S\textbf{L}R  & 68.95 & 73.10 & 64.85 & 54.89 & 60.04 & 49.56 & 59.33 & 59.21 & --& --\\
\bottomrule
\end{tabular}
\label{table:comprehension_automatic}
\end{table*}

Our experiments are designed to probe the behavior of models on VQD compared to RER datasets. To facilitate this, we created a variant of our VQDv1 dataset that had only a single correct bounding box.

To evaluate performance for the RER and `1 Obj' version of the VQDv1 dataset, systems only output a single bounding box during test time, so the \textbf{Precision@1} metric is used. For the `0-N Obj' version of the VQDv1 dataset, we use the standard PASCAL VOC metric \textbf{AP$^{IoU=.50}$} from object detection, which calculates the average precision across the dataset using an intersection over union (IoU) greater than 0.5 criteria for matching with the ground truth boxes.

\subsection{Models Evaluated}
We implemented and evaluated four models for VQD. All models are built on top of Faster R-CNN with a ResNet-101 backbone whose output bounding boxes pass through Non-Maximal Suppression with a threshold of 0.7. This acts as a region proposal generator that provides CNN features for each region.  

The four models we evaluate are:
\begin{enumerate}[nolistsep,noitemsep]
    \item \textbf{DETECT:} A model that uses the full Faster R-CNN system to detect all trained COCO classes, and then outputs the boxes that have the same label as the first noun in the query.
	\item \textbf{RANDOM:} Select one random Faster R-CNN proposal. 
    \item \textbf{Query-Blind:} A vision only model that does binary classification of each region proposal's CNN features using a 3 layer MultiLayer Perceptron(MLP) with 1024-unit hidden ReLU layers. 

    \item \textbf{Vision+Query (V+Q):} A model that does binary classification of each region proposal. The query features are obtained from the last hidden layer of a Gated Recurrent Unit (GRU) network, and then they are concatenated with the CNN features and fed into a 3 layer MLP with 1024-unit hidden ReLU layers.

\end{enumerate}

The primary reason for providing VQDv1 (1 obj.) and the RER results is to put the benefits of the VQD task in context. To aid in this endeavor, we also include comparison results directly from the SLR models~\cite{yu2017joint} for RER, which is a recent system for that task.

\subsection{Training Details}

The Query-Blind and Vision+Query models are trained with binary cross-entropy loss. We use a learning rate of 0.0004, and perform learning rate decay of 0.8 when the training loss plateaus continuously for five epochs. The best model is selected based on the validation loss after training for 50 epochs.

\subsection{Results}

Our main results are given in Table~\ref{table:comprehension_automatic}. Although simple, our Vision+Query model performs well across RER datasets, but it can also be applied to VQD tasks. As expected, RANDOM performs poorly on both VQDv1 datasets. DETECT beats RANDOM in the single object VQD setting by a large margin. Since, most of the questions in the RER datasets ask about common COCO categories, choosing one of those objects might be enough to get decent performance; however, DETECT performs poorly when evaluated under 0-$N$ object settings in VQDv1. To handle queries in VQD, models must be able to understand the context and comprehend multiple objects in isolation.

\section{Conclusion}
In this paper, we described our VQDv1 dataset as a test for visual grounding via goal-directed object detection. VQDv1 has both simple object presence and complex questions with $0$--$N$ bounding boxes. While VQDv1 contains only synthetically generated questions, this can help mitigate some forms of bias present in other VQA and RER datasets~\cite{cirik2018visual,kafle2017-cviu}. While it would be expensive, a large, carefully filtered, and well designed human annotated VQD dataset is the next step toward advancing visual grounding research. 

Compared to VQA, we argue that it is harder to be right for the \emph{wrong} reasons in VQD because methods must generate bounding boxes. Compared to RER, we argue that it is harder to exploit bias in VQD since there are a variable number of boxes per image, making it considerably more difficult, as demonstrated by our experiments. We believe the VQD approach has considerable value and can be used to advance visual grounding research.

\section*{Acknowledgments}  This work was supported in part by a gift from Adobe Research. The lab thanks NVIDIA for the donation of a GPU. We also thank fellow lab members Kushal Kafle and Tyler Hayes for their comments and useful discussions.

\bibliography{naaclhlt2019}
\bibliographystyle{acl_natbib}
\end{document}